\pdfoutput=1

\documentclass[11pt]{article}

\usepackage[]{naacl2021}

\usepackage{times}
\usepackage{latexsym}
\usepackage{breqn}
\usepackage{xspace}
\usepackage[normalem]{ulem}
\usepackage{booktabs}
\usepackage{multirow}

\definecolor{Economic}{rgb}{0.99, 0.4, 0.5}
\definecolor{Capacity}{rgb}{0.29, 0.59, 0.82}
\definecolor{Morality}{rgb}{0.0, 0.42, 0.24}
\definecolor{Fairness}{rgb}{1.0, 0.7, 0.0}
\definecolor{Legality}{rgb}{0.59, 0.44, 0.84}
\definecolor{Policy}{rgb}{0.7,1,1}
\definecolor{Crime}{rgb}{0.63, 0.79, 0.95}
\definecolor{Security}{rgb}{1,0.6,1}
\definecolor{Health}{rgb}{1, 0.9, 0.5}
\definecolor{Life}{rgb}{0.8, 0.3, 0.0}
\definecolor{Cultural}{rgb}{0.0, 0.1, 0.5}
\definecolor{Public}{rgb}{0.0, 0.66, 0.47}
\definecolor{Political}{rgb}{0.92, 0.3, 0.26}
\definecolor{External}{rgb}{0.94, 0.86, 0.6}
\definecolor{Other}{rgb}{1,0.4,1}

\DeclareRobustCommand{\legendsquare}[1]{%
  \textcolor{#1}{\rule{2ex}{1.5ex}}%
}

\newcommand{\friss}{{\sc Friss}\xspace}

\newcommand{\argzero}{\textsc{Arg0}\xspace}
\newcommand{\argone}{\textsc{Arg1}\xspace}

\newcommand{\argzeromm}{\operatorname{a}_0}
\newcommand{\argonemm}{\operatorname{a}_1}
\newcommand{\predmm}{\operatorname{p}}

\newcommand{\frissbox}[1]{{\setlength{\fboxsep}{1pt}{#1}}}

\usepackage[T1]{fontenc}

\usepackage[utf8]{inputenc}

\usepackage{microtype}

\title{Instructions for NAACL-HLT 2021 Proceedings}

\author{Shima Khanehzar$^{\dagger}$ \ \ Trevor Cohn$^{\dagger}$ \ \ Gosia Mikolajczak$^\text{\textborn}$ \ \ Andrew Turpin$^{\dagger}$ \ \ Lea Frermann$^{\dagger}$ \\
    $^{\dagger}$ School of Computing and Information Systems \\ $^\text{\textborn}$School of Social and Political Sciences \\ The University of Melbourne \\
    \texttt{skhanehzar@student.unimelb.edu.au} \\ \texttt{\{tcohn,mmikolajcza,aturpin,lfrermann\}@unimelb.edu.au}}

\begin{document}
\title{Framing Unpacked: A Semi-Supervised Interpretable Multi-View Model of Media Frames}
\maketitle
\begin{abstract}

Understanding how news media frame political issues is important due to its impact on public attitudes, yet hard to automate. Computational approaches have largely focused on classifying the frame of a full news article while framing signals are often subtle and local. Furthermore, automatic news analysis is a sensitive domain, and existing classifiers lack transparency in their predictions. This paper addresses both issues with a novel semi-supervised model, which jointly learns to embed local information about the events and related actors in a news article through an auto-encoding framework, and to leverage this signal for document-level frame classification. Our experiments show that: our model outperforms previous models of frame prediction; we can further improve performance with unlabeled training data leveraging the semi-supervised nature of our model; and the learnt event and actor embeddings intuitively corroborate the document-level predictions, providing a nuanced and interpretable article frame representation.

\end{abstract}

\section{Introduction}


Journalists often aim to package complex real-world events into comprehensive narratives, following a logical sequence of events involving a limited set of actors. Constrained by word limits, they necessarily select some facts over others, and make certain perspectives more salient. This phenomenon of {\it framing}, be it purposeful or unconscious, has been thoroughly studied in the social and political sciences~\cite{chong2007framing}. More recently, the natural language processing community has taken an interest in automatically predicting the frames of news articles \cite{DBLP:conf/emnlp/CardGBS16, DBLP:conf/emnlp/FieldKWPJT18, akyurek-etal-2020-multi, khanehzar-etal-2019-modeling, DBLP:conf/conll/LiuGMBW19, huguet-cabot-etal-2020-pragmatics}.

Definitions of framing vary widely including: expressing the same semantics in different forms (equivalence framing);  presenting selective facts and aspects (emphasis framing); and using established syntactic and narrative structures to convey information (story framing)~\cite{hallahan1999seven}. The model presented in this work builds on the concepts of emphasis framing and story framing, predicting the global (aka. primary) frame of a news article on the basis of the events and participants it features.

Primary frame prediction has attracted substantial interest recently with the most accurate models being supervised classifiers built on top of large pre-trained language models \cite{khanehzar-etal-2019-modeling, huguet-cabot-etal-2020-pragmatics}. This work advances prior work in two ways. First, we explicitly incorporate a formalization of {\it story framing} into our frame prediction models. By explicitly modeling news stories as latent representations over events and related actors, we obtain interpretable, latent representations lending transparency to our frame prediction models. We argue that transparent machine learning is imperative in a potentially sensitive domain like automatic news analysis, and show that the local, latent labels inferred by our model lend explanatory power to its frame predictions.

Secondly, the latent representations are induced without frame-level supervision, requiring only a pre-trained, off-the-shelf semantic role labeling (SRL) model \cite{DBLP:journals/corr/abs-1904-05255}. This renders our frame prediction models semi-supervised, allowing us to use large unlabeled news corpora.

More technically, we adopt a dictionary learning  framework  with  deep  autoencoders through which we learn to map events and their agents and patients\footnote{We experiment with three types of semantic roles: predicates and associated arguments (\argzero and \argone), however, our framework is agnostic to the types of semantic roles, and can further incorporate other types of semantic roles or labels.} independently into their respective structured latent space. Our model thus learns a latent multi-view representation of news stories, with each view contributing evidence to the primary frame prediction from its own perspective. We incorporate the latent multi-view representation into a transformer-based document-level frame classification model to form a semi-supervised model, in which the latent representations are jointly learnt with the classifier.

We demonstrate empirically that our semi-supervised model outperforms current state-of-the-art models in frame prediction. More importantly, through detailed qualitative analysis, we show how our latent features mapped to events and related actors allow for a nuanced analysis and add interpretability to the model predictions\footnote{Source code of our model is available at \href{https://github.com/shinyemimalef/FRISS}{https://github.com/shinyemimalef/FRISS}}. In summary, our contributions are:

\begin{itemize}
\item Based on the concepts of story- and emphasis framing, we develop a novel semi-supervised framework which incorporates local information about core events and actors in news articles into a frame classification model. 

\item We empirically show that our model, which incorporates the latent multi-view semantic role representations, outperforms existing frame classification models, with only labeled articles. By harnessing large sets of unlabeled in-domain data, our model can further improve its performance and achieves new state-of-the-art performance on the frame prediction task.

\item Through qualitative analysis, we demonstrate that the latent, multi-view representations 
aid interpretability of the predicted frames.
\end{itemize}

\section{Background and Related Work}
\label{sec:background}
A widely accepted definition of frames describes them as a selection of aspects of perceived reality, which are made salient in a communicating context to promote a particular problem definition, causal interpretation, moral evaluation and the treatment recommendation for the described issue \cite{entman1993framing}. While detecting media frames has attracted much attention and spawned a variety of methods, it poses several challenges for automatic prediction due to its vagueness and complexity. 

Two common approaches in the study of frames focus either on the detailed issue-specific elements of a frame or, somewhat less nuanced, on generic framing themes prevalent across issues. Within the first approach, \citet{matthes2008content} developed a manual coding scheme, relying on Entman's definition \cite{entman1993framing}. While the scheme assumes that each frame is composed of common elements, categories within those elements are often specific to the particular issue being discussed (e.g., ``same sex marriage'' or ``gun control''), making comparison across different issues, and detecting them automatically difficult. Similarly, earlier studies focusing specifically on unsupervised models to extract frames, usually employed topic modeling \cite{boydstun2013identifying, nguyen2015guided, DBLP:conf/acl/TsurCL15} to find the issue-specific frames, limiting across-issue comparisons.

Studies employing generic frames address this shortcoming by proposing common categories applicable to different issues. For example, \citet{boydstun2013identifying} proposed a list of 15~broad frame categories commonly used when discussing different policy issues, and in different communication contexts. The Media Frames Corpus (MFC; \newcite{DBLP:conf/acl/CardBGRS15}) includes about 12,000 news articles from 13 U.S. newspapers covering five different policy issues, annotated with the dominant frame from \citet{boydstun2013identifying}. Table \ref{table:frame dimension} in the Appendix lists all 15 frame types present in the MFC. The MFC has been previously used for training and testing frame classification models. \citet{DBLP:conf/emnlp/CardGBS16} provide an unsupervised model that clusters articles with similar collections of ``personas'' (i.e., characterisations of entities) and demonstrate that these personas can help predict the coarse-grained frames annotated in the MFC. While conceptually related to our approach, their work adopts the Bayesian modelling paradigm, and does not leverage the power of deep learning. \citet{DBLP:conf/acl/JiS17} proposed a supervised neural approach incorporating discourse structure. The current best result for predicting the dominant frame of each article in the MFC comes from \citet{khanehzar-etal-2019-modeling}, who investigated the effectiveness of a variety of pre-trained language models (XLNet, Bert and Roberta).

\begin{figure*}[t]
\centering
\includegraphics[width=0.99\textwidth,clip,trim=30 180 240 90]{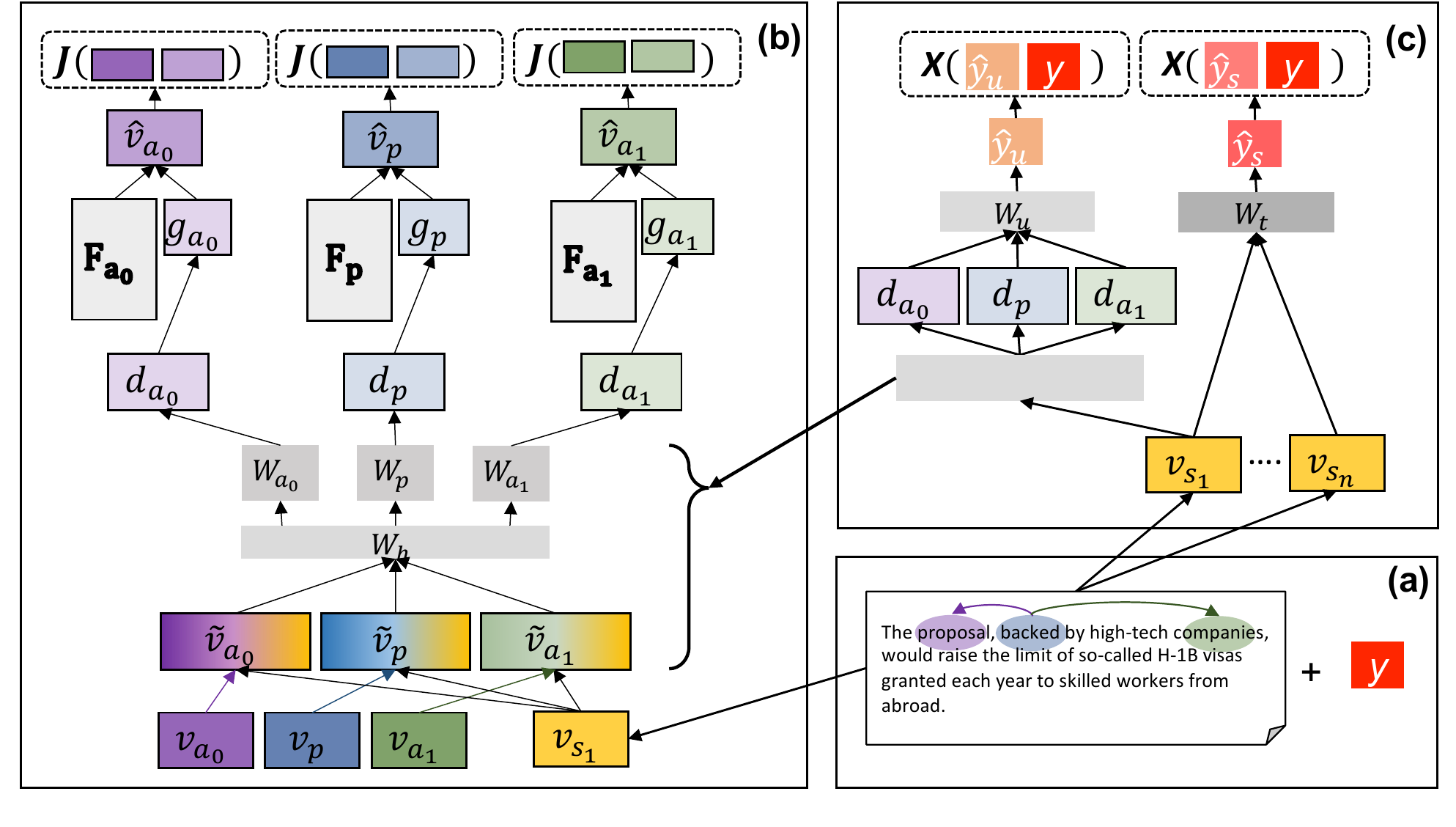} 

\caption{Our \friss model. {\bf (a)} Given an input document with a frame label~$y$ , we perform semantic role labeling (SRL) to obtain the predicate (blue), \argzero (purple) and \argone (green) for input sentences $s_1\ldots s_n$. {\bf (b)}~The unsupervised module takes as input semantic role embeddings ($v_{a_0},v_p,v_{a_1}$) and a sentence embedding ($v_{s_i}$), and learns latent, role-specific embedding matrices ($\mathbf{F}$) in an auto-encoding framework. The latent representations are incorporated into the overall frame classification module. {\bf(c)} We predict document-level frames based on transformer-based document embeddings ($\hat{y}_s$) and the view-specific latent representations ($\hat{y}_u$), using cross-entropy loss with the true frame label $y$.}
 \label{fig:model}
\end{figure*}

Recent methods have been expanded to multilingual frame detection. \citet{DBLP:conf/emnlp/FieldKWPJT18} used the MFC to investigate framing in Russian news. They introduced embedding-based methods for projecting frames of one language into another (i.e., English to Russian). \citet{akyurek-etal-2020-multi} studied multilingual transfer learning to detect multiple frames in target languages with few or no annotations. Recently,  \citet{huguet-cabot-etal-2020-pragmatics} investigated joint models incorporating metaphor, emotion and political rhetoric within multi-task learning to predict framing of policy issues.

Our modelling approach is inspired by recent advances in learning interpretable latent representations of the participants and relationships in fiction stories. \newcite{iyyer-etal-2016-feuding} present Relationship Modelling Networks (RMNs), which induce latent descriptors of types of {\it relationships} between characters in fiction stories, in an unsupervised way. RMNs combine dictionary learning with deep autoencoders, and are trained to effectively encode text passages as linear combinations over latent descriptors, each of which corresponds to a distinct relationship (not unlike topics in a topic model).  \newcite{DBLP:conf/emnlp/FrermannS17} extend the idea to a multi-view setup, jointly learning multiple dictionaries, which capture {\it properties} of individual characters in addition to relationships. We adopt this methodology for modeling news articles through three latent views: capturing their events (predicates), and participants (\argzero, \argone). We combine the unsupervised autoencoder with a frame classifier into an interpretable, semi-supervised framework for article-level frame prediction.



\section{Semi-supervised Interpretable Frame Classification} 
\label{model_overview}
In this section, we present our \texttt{Fr}ame classifier, which is \texttt{I}nterpretable and \texttt{S}emi-\texttt{s}upervised (\friss). The full model is visualized in Figure~\ref{fig:model}. Given a corpus of news articles, some of which have a label indicating their primary frame~$y$~ (Figure~\ref{fig:model}(a)), \friss learns to predict $\hat{y}$ for each document by combining a supervised classification module (Figure~\ref{fig:model}(c)) and an unsupervised\footnote{The autoencoder is unsupervised wrt. frame-level information, but it relies on an off-the-shelf semantic role labeler.} auto-encoding module (Figure~\ref{fig:model}(b)), which are jointly trained. The unsupervised module (i)~ can be trained with additional unlabeled training data, which improves performance (Section~\ref{ssec:exp2}); and (ii)~learns interpretable latent representations which improve the interpretability of
the  model (Section~\ref{ssec:exp3}).

Intuitively, \friss predicts frames based on aggregated sentence representation (supervised module; Section~\ref{model_supervised}) as well as aggregated fine-grained latent representations capturing actors and events in the article (unsupervised module; \ref{model_unsupervised}). The unsupervised module combines an auto-encoding objective with a multi-view dictionary learning framework~\cite{iyyer-etal-2016-feuding,DBLP:conf/emnlp/FrermannS17}. We treat predicates, their \argzero and \argone as three separate views, and learn to map each view to an individual latent space representative of their relation to the overall framing objective. Below, we will sometimes refer to views collectively as $z \in \{p, \argzeromm, \argonemm\}$. We finally aggregate the view-level representations and sentence representations to predict a document-level frame. The following sections describe \friss in technical detail.

\subsection{Unsupervised Module}
\label{model_unsupervised}

\subsubsection{Input} Each input document is sentence-segmented and automatically annotated by an off-the shelf transformer-based semantic role labeling model~\cite{DBLP:journals/corr/abs-1904-05255, DBLP:conf/conll/PradhanMXNBUZZ13} to indicate spans over the three semantic roles: predicates, \argzero{}s and \argone{}s.

We compute a contextualized vector representation for each semantic role span ($s_{\predmm}$, $s_{\argzeromm}$, $s_{\argonemm}$). We describe the process for obtaining predicate input representations $\boldsymbol{v}_p$ here for illustration. Contextualized representations for views $\argzeromm$ ($\boldsymbol{v}_{\argzeromm}$) and $\argonemm$ ($\boldsymbol{v}_{\argonemm}$) are obtained analogously. First, we pass each sentence through a sentence encoder, and obtain the predicate embedding by averaging all contextualized token representations $\boldsymbol{v}_w$ (of dimension $D_w$) in its span $s_p$ of length~$|s_p|$:

\begin{dmath}\label{eqn:span}
$$ {\boldsymbol{v}}_{\predmm} = \frac{1}{|s_p|} \sum_{w \in{s_p}}\boldsymbol{v}_w 
$$.
\end{dmath}

We concatenate $\boldsymbol{v}_{\predmm}$ with an overall sentence representation $\boldsymbol{v}_{s}$, which is computed by averaging all contextualized token embeddings of the sentence $s$ of length~$|s|$,\footnote{We also experimented with representing the sentence as the $[CLS]$ token embedding, but found it to perform worse empirically.}
\begin{align}
\label{eqn:vs}
 \boldsymbol{v}_{s} &= \frac{1}{|s|} \sum_{w \in{s}}\boldsymbol{v}_w \\
   \tilde{\boldsymbol{v}}_{\predmm} &= [{\boldsymbol{v}_{\predmm}}; \boldsymbol{v}_{s}],
 \end{align}
where [;] denotes vector concatenation. If a sentence has more than one predicate, a separate representation is computed for each of them.

\subsubsection{Multi-view Frame representations}
\label{ssec:dictmodel}

We combine ideas from auto-encoding (AE) and dictionary learning, as previously used to capture the content of fictitious stories~\cite{iyyer-etal-2016-feuding}, and its multi-view extension~\cite{DBLP:conf/emnlp/FrermannS17}. We posit a latent space as three view-specific dictionaries (Figure~\ref{fig:model} (b)) capturing events (predicates; $\boldsymbol{F}_p$), their first (\argzero; $\boldsymbol{F}_{\argzeromm}$) and second (\argone; $\boldsymbol{F}_{\argonemm}$) arguments, respectively. Given a view-specific input as described above, the autoencoder maps it to a low-dimensional distribution over ``dictionary terms'' (henceforth descriptors), which are learnt during training. The descriptors are vector-valued latent variables that live in word embedding space, and are hence interpretable through their nearest neighbors (~Table~\ref{tab:descriptors} shows examples of descriptors inferred by our model). By jointly learning the descriptors with the supervised classification objective, each descriptor will capture coherent information corresponding to a frame label in our supervised data set. We hence set the number of descriptors for each dictionary to $K=15$, the number of frames in our data set. For each view $z\in\{p,\argzeromm,\argonemm\}$, we define a dictionary $\boldsymbol{F}_z$ of dimensions $K \times D_w$.

More technically, our model follows two steps. First, we {\it encode} the input $\tilde{\boldsymbol{v}}_z$ of a known view $z$ by passing it through a feed forward layer $\boldsymbol{W}_h$ of dimensions $ 2D_w \times D_h$, shared across all the views, followed by a ReLU non-lineararity, and then another feed forward layer $\boldsymbol{W}_z$ of dimensions $D_h \times K$, specific to each view $z$. This results in a $K$-dimensional vector over the view-specific descriptors,
\begin{equation}
\label{e:h}
 \boldsymbol{l}_z = \mathbf{W}_z  \operatorname{ReLU}({\boldsymbol{W}_h} \tilde{\boldsymbol{v}}_z),
\end{equation}


Second, we reconstruct the original view embedding $\boldsymbol{v}_z$ as a linear combination of descriptors. While previous work used $\boldsymbol{l}_z$ directly as weight vector, we hypothesize that on our fine-grained semantic role level, only one or a few descriptors will be relevant to any specific span. We enforce this intuition using Gumbel-Softmax differentiable sampling with temperature annealing~\cite{DBLP:conf/iclr/JangGP17}. This allows us to gradually constrain the number of relevant descriptors used for reconstruction. We first normalize~$\boldsymbol{l}_z$,
\begin{equation}
\label{eqn:l}
 \boldsymbol{d}_z = \operatorname{Softmax}(\boldsymbol{l}_z),
 \end{equation}
and then draw $\boldsymbol{g}$ from the Gumbel distribution,\label{Gumbel distribution} and add it to our normalized logits $\boldsymbol{d_z}$ scaled by temperature $\tau$ , which is gradually annealed over the training phase:

\begin{equation}
 \begin{aligned}
 \boldsymbol{g} &\sim \operatorname{Gumbel}(0,1)\\
 \boldsymbol{g}_z &= \frac{\exp(\log(\boldsymbol{d}_z) + \frac{\boldsymbol{g}}{ 
\tau})}{\sum_f \exp(\log(\boldsymbol{d}_z) + \frac{\boldsymbol{g} }{\tau})}.
\end{aligned}
\end{equation}

We finally reconstruct the view-specific span embedding as

\begin{dmath}
\label{eqn-8}
$$ \hat{\boldsymbol{v}}_z = \boldsymbol{F}_z^T \boldsymbol{g}_z$$.
\end{dmath}

\subsubsection{Unsupervised Objective} 
\label{unsupervised_learning_objective}

\paragraph{Contrastive Loss}
We use the contrastive max-margin objective function following previous works in dictionary learning \cite{iyyer-etal-2016-feuding, DBLP:conf/emnlp/FrermannS17, han-etal-2019-permanent}. We randomly sample a set of negative samples~($N_{-}$) with the same view as the current input from the mini-batch. The unregularized objective $J_z^u$ (Eq. \ref{eq:j_u}) is a hinge loss that minimizes the L2 norm\footnote{We empirically found that L2 norm outperforms the dot product, and cosine similarity.} between the reconstructed embedding~ $\hat{\boldsymbol{v}}_z$ and the true input's view-specific embedding $\boldsymbol{v}_z$, while simultaneously maximizing the L2 norm between $\hat{\boldsymbol{v}}_z$ and negative samples~$\boldsymbol{v}_z^{n}$: 

\begin{dmath}
  J^u_z(\theta) =\frac{1}{\lvert N_{-}\rvert}\sum_{\boldsymbol{v}_z^n\in{N_{-}}} \max(0, 1 + 
l_2(\hat{\boldsymbol{v}}_z, \boldsymbol{v}_z) - l_2(\hat{\boldsymbol{v}}_z, 
\boldsymbol{v}_z^{n})), \label{eq:j_u}
 \end{dmath}
where $\theta$ represents the model parameters, $\lvert N_{-} \rvert $ is the number of negative samples, and the margin value is set to 1.

\paragraph{Focal Triplet Loss}\label{Focal Triplet Loss} Preliminary studies (Section \ref{ssec::p_studies}) suggested that some descriptors (aka frames) are more similar to each other than others. We incorporate this intuition through a novel mechanism to move the descriptors that are least involved in the reconstruction proportionally further away from the most involved descriptor.

Concretely, we select $t$ descriptors in ${\mathbf F}_z$ with smallest weights in $\boldsymbol{g}_z$ as additional negative samples. We denote the indices of the selected $t$ smallest components in $\boldsymbol{g}_z$ 
\begin{equation}
 I = [i_1, i_2, \dots, i_t].
\end{equation}

We use $\boldsymbol{F}^t_z$ to denote the matrix ($t\times D_w$) with only those $t$ descriptors. We re-normalize the weights of the selected $t$ descriptors, and denote the renormalized weights vector as $\boldsymbol{g}^t_z = [{g}^{i_1}_z,{g}^{i_2}_z, \dots, {g}^{i_t}_z]$. For each element in $\boldsymbol{g}^t_z$, we compute an individual margin based on its magnitude. Intuitively, the smaller the weight is, the larger its required margin from a given total margin budget $\lvert M \rvert$,

\begin{dmath}
{m}_{z}^{i_t} = \lvert M \rvert * (1 - g_{z}^{i_t})^2.
\end{dmath}

We compute the standard margin-based hinge loss over the additional negative samples with sample-specific margins:

\begin{dmath}
J_z^t(\theta) =\frac{1}{\lvert T \rvert}\sum_{i_t \in I} \max(0, m_{z}^{i_t} + 
l_2(\hat{\boldsymbol{v}}_z, \boldsymbol{v}_z) - l_2(\hat{\boldsymbol{v}}_z, \boldsymbol{v}_z^{i_t})).
\end{dmath}

We sum the focal triplet objective $J_z^t$ with $J_z^u$, and then sum over all specific spans $s \in S_z$, while adding an additional orthogonality encouraging regularization term.
\begin{equation}
J_z(\theta) = \sum_{s \in S_z}(J^u_z + J^t_z) + \lambda||\bf{F_z}\bf{F^T_z} - 
\bf{I}||^2_F,
\end{equation}
where $\lambda$ is a hyper-parameter that can be tuned. We finally aggregate the loss from all the views:
\begin{equation}
\label{e:unsuploss}
J(\theta) = \sum_{z \in \{\predmm,\argzeromm,\argonemm\}} J_z(\theta).
\end{equation}

\subsection{Supervised Document-level Frame Classification} 
\label{model_supervised}
We incorporate the semantic role level predictions as described above into a document-level frame classifier consisting of two parts, which are jointly learnt with the unsupervised model described above: (i)  a classifier based on aggregated span-level representations computed as described in Sec~\ref{model_unsupervised} (Fig.~\ref{fig:model} (c; left); Sec.~\ref{ssec:spanclass}) and (ii) a classifier based on an aggregated sentence representations (Fig.~\ref{fig:model} (c; right); Sec.~\ref{ssec:classifier}).

\subsubsection{Span-based Classifier}
\label{ssec:spanclass}
The unsupervised module makes predictions on the semantic role span level, however, our goal is to predict document-level frame labels. We aggregate span-level representations $\boldsymbol{d}_z$ (Eq.~\ref{eqn:l}) by averaging across spans and then views:\footnote{We empirically found these representations to outperform the sparser $\boldsymbol{g}_z$.}
\begin{equation}
\label{eqn-15}
\begin{aligned}
\boldsymbol{w}_u &= \frac{1}{Z} \sum_{z\in\{\predmm,\argzeromm,\argonemm\}}\ 
\frac{1}{|S_z|}\sum_{s\in S_z} \boldsymbol{d}^s_z\\
\hat{y}_u &= \operatorname{Softmax}(\boldsymbol{w}_u),
\end{aligned}
\end{equation}
where $Z$ is the number of the views, and $S_z$ are the set of view-specific spans in the current document. We finally pass the logits through a softmax layer to predict a distribution over frames.

\subsubsection{Sentence-based Classifier}
\label{ssec:classifier}
We separately predict a document-level frame based on the aggregate sentence level representations computed in Eq.~(\ref{eqn:vs}). We first pass each sentence embedding through a feed forward layer $W_r$ of dimensions $D_w \times D_w$, followed by a ReLU non-linearity, and another feed forward layer $W_t$ to map the resulting representation to $K$ dimensions. Then average across sentences of the current document $S_d$ and pass the result through a softmax layer,
\begin{equation}
\label{eqn:ys}
\begin{aligned}
\boldsymbol{w}_s &= \operatorname{ReLU}(\boldsymbol{W}_{r} \boldsymbol{v}_s)\\
\hat{y}_s &= \operatorname{Softmax}\bigg(\frac{1}{\lvert S_d \rvert} \sum_{s \in S_d}\boldsymbol{W}_{t} 
\boldsymbol{w}_s\bigg).
\end{aligned}
\end{equation}

\subsection{Full Loss}
\label{ssec:suploss}
We jointly train the supervised and unsupervised model components. The supervised loss $X(\theta)$ consists of two parts, one for the sentence-based classification and one for the aggregated span-based classification:

\begin{equation}
 X(\theta) = X(\hat{y}_u, y) + X(\hat{y_s}, y).
\end{equation}
The full loss balances the supervised and unsupervised components with a 
hyper-parameter~$\alpha$:
\begin{equation}
\label{eqn-combined-loss}
 L(\theta) = \alpha \times X(\theta) + (1 - \alpha) \times J(\theta).
\end{equation}

\section{Experimental Settings}

\paragraph{Dataset}

We follow prior work on automatic prediction of a single, primary frame of a news article as annotated in the Media Frames Corpus (MFC;~\newcite{DBLP:conf/acl/CardBGRS15}). The MFC contains a large number of news articles on five contentious policy issues (immigration, smoking, gun control, death penalty, and same-sex marriage), manually annotated with document- and span-level frames labels from a set of 15 general frames (listed in Table \ref{table:frame dimension} in the Appendix). Articles were selected from 13 major U.S. newspapers, published between 1980 and 2012. Following previous work, we focus on the immigration portion of MFC, which comprises 5,933 annotated articles, as well as an additional 41,286 unlabeled articles. The resulting dataset contains all 15 frames. Table \ref{table:frame dimension} (Appendix) lists the corresponding frame distribution. We partition the labeled dataset into 10 folds, preserving the overall frame distribution for each fold.

\paragraph{Pre-processing and Semantic Role labeling}
We apply state-of-art BERT-based SRL model \cite{DBLP:journals/corr/abs-1904-05255} to obtain SRL spans for each sentence. The off-the-shelf model from AllenNLP is trained on OntoNotes5.0 (close to 50\% news text). While a domain-adapted model may lead to a small performance gain, the off-the-shelf model enhances generalizability and reproducibility. Qualitative examples of detected SRL spans are shown in Table~\ref{tab:qualdocs}, which confirm that SRL predictions are overall accurate.

We extract semantic role spans for predicates, their associated first (\argzero) and second (\argone) arguments for each sentence in a document. For the unsupervised component, we disregard sentences with no predicate, and sentences missing both \argzero and \argone.

\paragraph{Sentence Encoder}
In all our experiments, we use RoBERTa \cite{DBLP:journals/corr/abs-1907-11692} as our sentence encoder, as previous work \cite{khanehzar-etal-2019-modeling} has shown that it outperforms BERT \cite{devlin2019bert} and XLNet \cite{yang2019xlnet}. We pass each sentence through RoBERTa and retrieve the token-level embeddings. To obtain the sentence embedding, we average the RoBERTa embeddings of all words (Eq.~\ref{eqn:vs}). To obtain SRL span embeddings, we average the token embeddings of all words in a predicted span (Eq.~\ref{eqn:span}). Following \citet{gururangan-etal-2020-dont}, we pre-train RoBERTa  with immigration articles using the masked language model (MLM) objective. Only the labeled data is used for pre-training for fair comparison between \friss and previous models.

\begin{table}[t] 
{\small
\centering
\begin{tabular}{l c c}
\toprule 
\textbf{Model} & \textbf{Acc.} & \textbf{Macro-F1}\\
\midrule 
\citet{DBLP:conf/emnlp/CardGBS16}    & 56.8  & -\\
\citet{DBLP:conf/emnlp/FieldKWPJT18} & 57.3  & -\\
\citet{DBLP:conf/acl/JiS17}          & 58.4  & -\\
\citet{khanehzar-etal-2019-modeling} & {65.8} & -\\
\midrule 
RoBERTa-S & 66.4 (0.008)& 58.1 (0.013)\\
RoBERTa-S+MLM & 67.1 (0.008) & 58.8 (0.013)\\
\midrule 
FRISS (labeled only) & \textbf{68.8} (0.009) & \textbf{60.1} (0.011) \\
FRISS (labeled+unlabeled) & \textbf{69.7} (0.011) & \textbf{60.5} (0.015) \\
\bottomrule 
\end{tabular}
}
\caption{\label{tab:supervised_result} Primary Frame prediction on the MFC immigration data set with mean (standard deviation) over 10-fold cross-validation. 
We compare our full model \friss against recent work, and the supervised RoBERTa component with and without pre-training. The results are statistically significant (p < 0.05; paired sample t-test). FRISS (labeled only) vs RoBERTa-S+MLM: p=0.009;
FRISS (labeled + all unlabeled) vs FRISS (labeled only): p=0.012;
FRISS (labeled + all unlabeled) vs RoBERTa-S+MLM: p=0.003.}
\end{table}

\paragraph{Parameter Settings}
We set the maximum sequence length to RoBERTa 64 tokens, the maximum number of sentences per document to 32, and the maximum number of predicates per sentence to 10.\footnote{$96\%$ of the sentences are under 64 tokens; $95\%$ of the documents have less than 32 sentences, and $\gg99\%$ of the sentences have less than 10 predicates.} We set the number of dictionary terms $K=15$, i.e.,~the number of frame classes in the MFC corpus. Each dictionary term is of dimension $D_w=768$, equal to the RoBERTa token embedding dimension. We also fix the dimensions of hidden vector $\boldsymbol{w}_s$ (Eqn.~\ref{eqn:ys}) and $D_h$ to this value. We set the number of descriptors in Focal Triplet Loss $t=8$ and the margin pool $\lvert M \rvert = t$. We set the balancing hyper-parameter between the supervised and unsupervised loss $\alpha=0.5$ , and $\lambda = 10^{-3}$. The dropout rate is set to $0.3$.

We perform stochastic gradient descent with mini-batches of 8 documents. We use the Adam optimizer \cite{DBLP:journals/corr/KingmaB14} with the default parameters, except for the learning rate, which we set to $2\times10^{-5}$ (for the RoBERTa parameters) and $5\times10^{-4}$ (for all other parameters). We use a linear scheduler for learning rate decay. The weight decay is applied to all parameters except for bias and batch normalization. We update the Gumbel softmax temperature with the schedule: $\tau = \max(0.5, \exp(-5\times10^{-4} \times \operatorname{iteration})$, updating the temperature every 50 iterations. For all our experiments, we run a maximum of 10 epochs, evaluate every $50$ iterations, and apply early-stopping if the accuracy does not improve for 20 consecutive evaluations.



\section{Evaluation}

In this section, we evaluate the performance of \friss on primary frame prediction for issue-specific news articles against prior work (Sec~\ref{ssec:exp1}), demonstrate the benefit of adding additional unlabeled data to our semi-supervised model (Sec~\ref{ssec:exp2}), and present a qualitative analysis of our model output corroborating its interpretability (Sec~\ref{ssec:exp3}).

\paragraph{Preliminary Studies} \label{ssec::p_studies} Preliminary analyses revealed that human annotators disagree on frame labels non-uniformly, suggesting that some pairs of frames are perceived to be more similar than others. This observation motivated the Focal Triplet Loss and Gumbel regularization components of our model. In particular, the following groups of frame labels are confused most frequently \{"Policy Prescription and Evaluation", "Public Sentiment", "Political"\}, \{"Fairness", "Legality"\}, \{"Crime and Punishment", "Security and Defense"\}, and \{"Morality", "Quality of Life", "Cultural Identity"\}. This ovservation is also corroborated through the empirical gain through the focal triplet loss (Table~\ref{supervised_ablation_result}).

\begin{table}
\centering
\begin{tabular}{l c c}
\toprule
\textbf{Model} & \textbf{Acc.} & \textbf{Macro-F1}\\
\midrule
\friss  & \textbf{68.83} &  \textbf{60.05} \\
- focal & 68.73 & 59.98 \\
- Gumbel & 68.51 & 59.70 \\
- focal, Gumbel & 68.47 & 59.72 \\
\midrule
$\predmm$ only & 68.35 & 59.50 \\
$\argzero$ only & 68.47 & 59.68 \\
$\argone$ only & 68.53 & 59.71 \\
\bottomrule
\end{tabular}
\caption{\label{supervised_ablation_result}
Ablation results for \friss on primary frame prediction. Top removing Focal Triplet Loss (focal; \ref{Focal Triplet Loss}), and/or Gumbel distribution (Gumbel; \ref{Gumbel distribution}). Bottom \friss trained with only a single view (predicate, \argzero and \argone) on primary frame prediction.}
\end{table}
\subsection{Experiment 1: Frame Prediction}
\label{ssec:exp1}
For the supervised model, we report accuracy, as has been done in previous work, as well as Macro-F1, which is oblivious to class sizes, shedding light on performance across all frames. Table~\ref{tab:supervised_result} compares \friss against related work. \newcite{DBLP:conf/emnlp/CardGBS16} incorporate latent personas learnt with a Bayesian model; \newcite{DBLP:conf/emnlp/FieldKWPJT18} derive frame-specific lexicons based on pointwise-mutual information; \newcite{DBLP:conf/acl/JiS17} incorporate a supervised discourse classifier, and~\newcite{khanehzar-etal-2019-modeling} train frame classifiers on top of RoBERTa-based document embeddings. RoBERTa-S corresponds to the sentence-embedding based component of \friss (Fig~\ref{fig:model}(b); left) without and with (+MLM) unsupervised pre-training. Overall, we can see that all our model variants outperform previous work in terms of both accuracy and macro-F1. Experiments were run 5 times with 10-fold cross-validation. The results in Table~\ref{tab:supervised_result} are statistically significant (p<0.05; paired sample t-test).

To better understand the impact of our various model components,
we performed an ablation study on Focal Triplet Loss, the Gumbel regularization, and the impact of individual views. Table~\ref{supervised_ablation_result} shows that both the focal loss and the Gumbel regularization contribute to model performance. Training \friss with any single view individually leads to a performance drop, which is most drastic if the two arguments are omitted, suggesting that the model relies on both predicate and argument information, with arguments playing a slightly more important role.

\subsection{Experiment 2: Benefit of Unlabeled Data}
\label{ssec:exp2}
\begin{figure}[t]
 \includegraphics[width=0.48\textwidth]{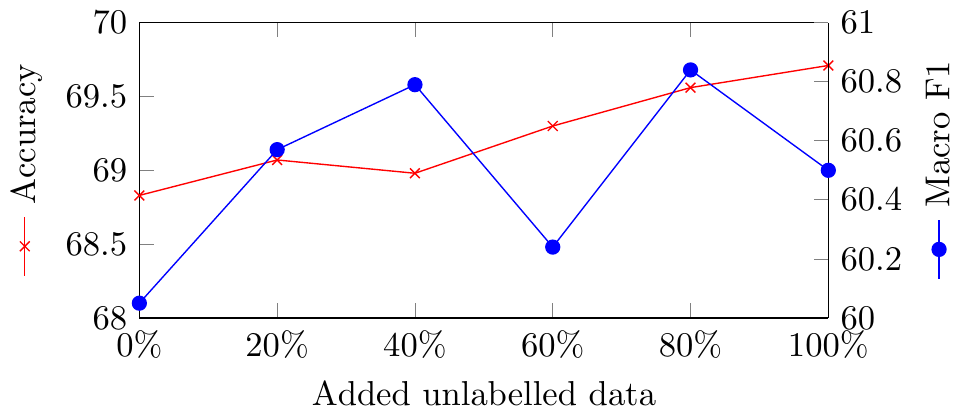}
\caption{\label{more_unlabeled_results}
\friss frame prediction performance with different portions of the 41K unlabeled documents.}
\end{figure}

Our semi-supervised model can leverage news articles without a frame label, in addition to a labeled training set. We investigated the impact of training \friss with different amounts of additional news articles, taken from the unlabeled immigration portion of the MFC. Figure~\ref{more_unlabeled_results} shows the impact of additional unlabeled data on accuracy and F1: Models with access to more unlabelled data tend to result in higher accuracy and Macro F1 scores. Given the abundance of online news articles, this motivates future work on minimally supervised frame prediction, minimizing the reliance on manual labels and maximizing generalizability to new issues, news outlets or languages.

\begin{table}[t]

{\setlength{\tabcolsep}{0.2em}
    \small{\begin{tabular}{ll}
\toprule
  \multirow{4}{*}{\rotatebox{90}{\argzero}} &
    \legendsquare{Capacity} USCIS, state department, agency, federal official \\ 
  & \legendsquare{Political} Trump, house republican, Obama, democrat, senate\\
  & \legendsquare{Legality} supreme court, justice, federal judge, court\\
    & \legendsquare{Public} organizer, activist, protester, demonstrator, marcher\\  \midrule

  \multirow{4}{*}{\rotatebox{90}{\textsc{Pred}}} &
  \legendsquare{Capacity} process, handle, swamp, accommodate, wait, exceed\\
  & \legendsquare{Political} veto, defeat, vote, win, introduce, endorse, elect\\
  & \legendsquare{Legality} sue, uphold, entitle, appeal, shall, violate, file\\
    & \legendsquare{Public} chant, march, protest, rally, wave, gather, organize\\\midrule
    
  \multirow{4}{*}{\rotatebox{90}{\argone}} &
  \legendsquare{Capacity} application, foreign worker, visa, applicant\\
  & \legendsquare{Political} amendment, reform, legislation, voter, senate bill\\
  & \legendsquare{Legality} political asylum, asylum, lawsuit, suit, status, case\\
    & \legendsquare{Public} rally, marcher, march, protest, movement, crowd\\
  \bottomrule
 \end{tabular}}}
 \caption{\label{tab:descriptors} Spans inferred as most highly associated with the Capacity \& Resources (\legendsquare{Capacity}), Political (\legendsquare{Political}), Legality (\legendsquare{Legality}), and Public Sentiment (\legendsquare{Public}) frames, for each view (\argzero, \textsc{Pred}, \argone).}
\end{table}

{
\setlength{\tabcolsep}{0.1em}
{
\renewcommand{\arraystretch}{0.0}
\begin{table*}[h!]
\footnotesize 
\centering
\begin{tabular}{|p{\textwidth}|}

\hline\vspace{1ex}
{\bf Frame: }
\legendsquare{Capacity} Capacity \& Resources \hspace{3ex}
\legendsquare{Political} Political \hspace{3ex}
\legendsquare{Legality} Legality \hspace{3ex}
\legendsquare{Public} Public Sentiment\vspace{1ex}\\\hline

\end{tabular}

\vspace{1ex}
\begin{tabular}{|p{\textwidth}|}
\hline
\frissbox{\colorbox{Legality!80}{BILL ON IMMIGRANT WORKERS}}$^{\argonemm}$ \frissbox{\colorbox{Political!65}{{DIES}}}$^{\predmm}$. \frissbox{\colorbox{Capacity!80}{{Legislation}}}$^{\argzeromm}$ to \frissbox{\colorbox{Capacity!100}{allow}}$^{\predmm}$ nearly twice as many \frissbox{\colorbox{Capacity!100}{computer-savvy foreigners and other high-skilled immigrants}}$^{\argonemm}$ into the country next year apparently has \frissbox{\colorbox{Capacity!60}{died}}$^{\predmm}$ in Congress. \frissbox{\colorbox{Political!95}{The House}}$^{\argzeromm}$ \frissbox{\colorbox{Political!95}{passed}}$^{\predmm}$ \frissbox{\colorbox{Political!95}{the compromise measure}}$^{\argonemm}$ last month, 288-133, but \frissbox{\colorbox{Political!95}{Sen. Tom Harkin, D-Iowa}}$^{\argzeromm}$, had \frissbox{\colorbox{Capacity!70}{blocked}}$^{\predmm}$ 
\colorbox{Political!100}{a vote}$^{\argonemm}$ when in the Senate. 
\frissbox{\colorbox{Capacity!80}{The proposal}}$^{\argzeromm, \argonemm}$, \colorbox{Capacity!80}{backed}$^{\predmm}$
\frissbox{\colorbox{Capacity!60}{by high-tech companies}}$^{\argzeromm}$, would
\frissbox{\colorbox{Capacity!50}{raise}}$^{\predmm}$ 
\frissbox{\colorbox{Capacity!100}{the limit of so-\frissbox{\colorbox{Capacity!90}{called}} H-1B visas}}$^{\argonemm}$
\frissbox{\colorbox{Capacity!100}{granted}}$^{\predmm}$ each year to skilled workers from abroad. 
\frissbox{\colorbox{Capacity!100}{Only 65,000 visas}}$^{\argonemm}$ are now \colorbox{Capacity!100}{granted}$^{\predmm}$ each year; 
\frissbox{\colorbox{Capacity!100}{the bill}}$^{\argzeromm}$ would 
\frissbox{\colorbox{Capacity!100}{raise}}$^{\predmm}$ 
\frissbox{\colorbox{Capacity!100}{the annual cap}}$^{\argonemm}$ to 115,500 for the next two years and to 107,500 in 2001. 
\frissbox{\colorbox{Capacity!100}{The ceiling}}$^{\argonemm}$ would 
\frissbox{\colorbox{Capacity!100}{return}}$^{\predmm}$ to 65,000 in 2002.\\\hline
\end{tabular}

\vspace{1ex}
\begin{tabular}{|p{\textwidth}|}
\hline
The Fix: Immigration all of a sudden a top campaign issue. 1. 
\frissbox{\colorbox{Legality!99}{The Obama administration's decision}}$^{\argzeromm}$ to move forward with a legal challenge to Arizona's stringent illegal immigration law will almost certainly 
\frissbox{\colorbox{Legality!90}{elevate}}$^{\predmm}$ 
\frissbox{\colorbox{Legality!90}{the issue on the campaign trail}}$^{\argonemm}$ this fall. 
\frissbox{\colorbox{Political!100}{The Arizona measure}}$^{\argonemm}$, which was 
\frissbox{\colorbox{Political!95}{signed}}$^{\predmm}$ into law by 
\frissbox{\colorbox{Political!95}{Gov. Jan Brewer (R)}}$^{\argzeromm}$ in April, is a major political touchstone--of prime importance to Hispanics, the fastest 
\frissbox{\colorbox{Political!95}{growing}}$^{\predmm}$ 
\frissbox{\colorbox{Political!100}{demographic group}}$^{\argonemm}$ in the country and a coveted electoral prize for both parties.
\frissbox{\colorbox{Political!100}{Democratic strategists}}$^{\argzeromm}$
\frissbox{\colorbox{Political!100}{see}}$^{\predmm}$ 
\frissbox{\colorbox{Political!100}{the Arizona law}}$^{\argonemm}$ as a key moment in the ongoing battle to 
\frissbox{\colorbox{Political!95}{win}}$^{\predmm}$ 
\frissbox{\colorbox{Political!100}{the loyalty of Hispanic voters}}$^{\argonemm}$. 
\frissbox{\colorbox{Political!100}{They}}$^{\argzeromm}$ 
\frissbox{\colorbox{Political!95}{believe}}$^{\predmm}$ that 
\frissbox{\colorbox{Political!95}{it}}$^{\argonemm}$ will have a similar chilling effect for Republicans with Latinos as the passage of California's Proposition 187 did in the 1990s. 
\frissbox{\colorbox{Political!100}{Republicans}}$^{\argzeromm}$, on the other hand, 
\frissbox{\colorbox{Political!95}{believe}}$^{\predmm}$ 
\frissbox{\colorbox{Political!95}{that Democrats are badly out of step with the American people on the immigration issue}}$^{\argonemm}$.
\frissbox{\colorbox{Political!75}{They}}$^{\argzeromm}$ 
\frissbox{\colorbox{Political!80}{cite}}$^{\predmm}$ 
\frissbox{\colorbox{Political!70}{the Obama administration's aggressive approach}}$^{\argonemm}$ to 
\frissbox{\colorbox{Legality!40}{fighting}}$^{\predmm}$ 
\frissbox{\colorbox{Legality!60}{the Arizona law}}$^{\argonemm}$ is yet more evidence of that out-of-touchness. In that vein,
\frissbox{\colorbox{Political!95}{nearly two dozen House Republicans}}$^{\argzeromm}$ 
\frissbox{\colorbox{Political!95}{sent}}$^{\predmm}$ 
\frissbox{\colorbox{Political!50}{\frissbox{\colorbox{Legality!90}{a letter}}}}$^{\leftarrow \argonemm, \argzeromm\rightarrow}$ to Attorney General Eric Holder on Tuesday
\frissbox{\colorbox{Political!40}{describing}}$^{\predmm}$ 
\frissbox{\colorbox{Legality!30}{the legal challenge}}$^{\argonemm}$ as the "height of irresponsibility and arrogance."
\frissbox{\colorbox{Public!80}{\frissbox{\colorbox{Public!50}{Polling}$^{\predmm}$ on the Arizona law}}}$^{\argonemm}$  specifically 
\frissbox{\colorbox{Public!80}{falls}}$^{\predmm}$ in Republicans' favor, although 
\frissbox{\colorbox{Public!80}{broader data}}$^{\argzeromm}$  
\frissbox{\colorbox{Public!70}{suggests}}$^{\predmm}$ 
\frissbox{\colorbox{Public!80}{\frissbox{\colorbox{Public!80}{a public}}}$^{\argonemm}$ deeply 
\frissbox{\colorbox{Public!80}{divided}$^{\predmm}$ on immigration}}. In the latest Washington Post/ABC poll, 
\frissbox{\colorbox{Public!80}{58 percent}}$^{\argzeromm}$  
\frissbox{\colorbox{Public!80}{expressed}}$^{\predmm}$ 
\frissbox{\colorbox{Public!80}{support for the Arizona law}}$^{\argonemm}$ -- 
\frissbox{\colorbox{Public!80}{including}}$^{\predmm}$ 
\frissbox{\colorbox{Public!80}{42 percent who were strongly supportive}}$^{\argonemm}$ -- while 
\frissbox{\colorbox{Public!100}{41 percent}}$^{\argzeromm}$  
\frissbox{\colorbox{Public!80}{opposed}}$^{\predmm}$ 
\frissbox{\colorbox{Public!80}{it}}$^{\argonemm}$. \\\hline
\end{tabular}
\caption{\label{tab:qualdocs}Two articles from the MFC, annotated with SRL span-level frame predictions generated by \friss. The true frame label of both articles is Political (red). Each detected span ($\predmm$, $\argzeromm$ or $\argonemm$) has been highlighted with its most closely associated frame. Darker shades indicate higher confidence. The top document is mis-classified as ``Capacity \& Resources'' (blue), the bottom document is classified correctly.}
\end{table*}
}}

\subsection{Experiment 3: Qualitative Evaluation}
\label{ssec:exp3}

In this experiment, we explore the added interpretability contributed by the local latent frame representations. Table~\ref{tab:qualdocs} contains two MFC documents, 
highlighted with the most highly associated frame for each identified 
span for $\predmm$, $\argzeromm$ or $\argonemm$. We can observe that the frame associations (a)~are intuitively meaningful; and (b)~provide a detailed account of the predicted primary frame. For both documents the gold primary frame is `Political', the bottom document is classified correctly, whereas the top document is mis-classified as `Capacity \& Resources'. The detailed span-level predictions help to explain the model prediction, and in fact add support for the the mis-prediction, suggesting that predicting a single primary document frame may be inappropriate. In the bottom document ``a letter'' serves as both $\argonemm$ of ``Republicans sent a letter'', where it is predicted as `Political', and as $\argzeromm$ of the clause ``a letter [...] describing the legal challenges'', where it is classified as `Legality', another example of the nuance of our model predictions, which can support further in-depth study of issue-specific framing.

The potential of our model for fine-grained frame analysis is illustrated in Table~\ref{tab:qualdocs}, which shows how each particular SRL span contributes differently towards various frame categories. It adds a finer-grained framing picture, and estimate of the trustworthiness of model predictions. It allows to assess the main actors wrt. a particular frame (within and across articles), as well as the secondary frames in each article. Also, using SRL makes our model independent of human annotation, and more generalizable. Going beyond “highlighting indicative phrases”, our model can distinguish their roles (e.g., the “ICE” as an actor vs. participant in a particular frame).

Table~\ref{tab:descriptors} shows the semantic role spans, which are most closely related to Capacity \& Resources (blue), Political (red), Legality (purple) and Public Sentiment (green) descriptors in the latent space. We can observe that all associated spans are intuitively relevant to the \{frame, view\}. Furthermore, \argzero spans tend to correspond to active participants (agents) in the policy process (including politicians and government bodies), whereas \argone spans illustrate the affected participants (patients such as foreign workers, applicants), processes (reforms, cases, movements), or concepts under debate (political asylum). In future work, we aim to leverage these representations in scalable, in-depth analyses of issue-specific media framing. A full table illustrating the learnt descriptors for all 15~frames in the MFC and all three views is included in Table \ref{tab:descriptors_supplementary} in Appendix.


\section{Conclusion}
We presented \friss, an interpretable model of media frame prediction, incorporating notions of emphasis framing (selective highlighting of issue aspects) and story framing (drawing on the events and actors described in an article). Our semi-supervised model predicts article-level frame of news articles, leveraging local predicate and argument level embeddings. We demonstrated its three-fold advantage: first, our model empirically outperforms existing models for frame classification; second, it can effectively leverage additional unlabeled data further improving performance; and, finally, its latent representations add transparency to classifier predictions and provide a nuanced article representation. The analyses provided by our model can support downstream applications such as automatic, yet transparent, highlighting of reporting patterns across countries or news outlets; or frame-guided summarization which can support both frame-balanced or frame-specific news summaries. In future work, we plan to extend our work to more diverse news outlets and policy issues, and explore richer latent models of article content, including graph representations over all involved events and actors.

\section*{Acknowledgments}
We thank the anonymous reviewers for their helpful feedback and suggestions. This article was written with the support from the graduate research scholarship from the Melbourne School of Engineering, University of Melbourne provided to the first author. The original news articles used in this work were obtained from Lexis Nexis under the institutional licence held by the University of Melbourne. This research was undertaken using the LIEF HPC-GPGPU Facility hosted at the University of Melbourne, established with the assistance of LIEF Grant LE170100200. 

\bibliography{bibfile}
\bibliographystyle{acl_natbib}

\appendix
\begin{table*}
\begin{tabular}{lp{7.5cm}l}
\toprule
{\bf Frame} & {\bf Frame description} & {\bf \% IMM}\\
\midrule
\textbf{Economic}: &costs, benefits, or other financial implications & 7.0\% \\
\textbf{Capacity and Resources}: &availability of physical, human or financial resources, and capacity of current systems Morality: religious or ethical implications & 3.5\%\\
\textbf{Morality}: &religious or ethical implications & 1.3\%\\
\textbf{Fairness and Equality}:& balance or distribution of rights, responsibilities, and resources & 2.6\%\\
\textbf{Legality, Constitutionality, Jurisdiction}: &rights, freedoms, and authority of individuals, corporations, and government & 16.1\%\\
\textbf{Policy Prescription and Evaluation}: &discussion of specific policies aimed at addressing problems & 8.0\% \\
\textbf{Crime and Punishment}: &effectiveness and implications of laws and their enforcement & 13.6\%\\
\textbf{Security and Defence}: &threats to welfare of the individual, community, or nation & 4.9\%\\
\textbf{Health and Safety}:& health care, sanitation, public safety & 4.0\%\\
\textbf{Quality of Life}:& threats and opportunities for the individual’s wealth, happiness, and well-being & 7.0\%\\
\textbf{Cultural Identity}:& traditions, customs, or values of a social group in relation to a policy issue & 9.4\%\\
\textbf{Public Sentiment}:& attitudes and opinions of the general public, including polling and demographics & 4.1\%\\
\textbf{Political}: &considerations related to politics and politicians, including lobbying, elections, and attempts to sway voters & 16.3\% \\
\textbf{External Regulation and Reputation}: &international reputation or foreign policy of the U.S. & 2.2\%\\
\textbf{Other}: &any coherent group of frames not covered by the above categories & 0.2\%\\
\midrule
\end{tabular}
\caption{Framing dimensions from \cite{boydstun2013identifying}. The final column (\% IMM) denotes the frame prevalence in the Immigration portion of the MFC used in the experiments reported in this paper.}

\label{table:frame dimension}
\end{table*}

\section{Frame Descriptors Extractions}
We denote the set of spans that belong to a specific \{frame, view\} combination as $C^f_z$, for all possible combinations of frame categories and views. We obtain the high confidence set $\hat{C}^f_z$ for each specific \{frame, view\} combination by only preserving the spans whose $g_z^f > 0.8$. To obtain a more general sets of spans, we remove the stop-words and lemmatize the spans. To normalize the sets of high-confidence spans, we calculate the inverse-document frequency for each span $C^f_z$ and sort the spans accordingly by the inverse-document frequency to obtain the representative descriptors. Table \ref{tab:descriptors_supplementary} illustrating the learnt descriptors for all 15~frames in the MFC and all three views.

\begin{table*}

{\setlength{\tabcolsep}{0.3em}
    \small{\begin{tabular}{|ll|}
\toprule
   \multirow{14}{*}{\argzero} &
    
    \legendsquare{Capacity} USCIS, state department, agency, federal official, IN , immigration service, ASA international, visitor \\
    
    & \legendsquare{Political} Trump, house republican, Obama, democrat, senate, rubio, tancredo, Mr. romney, Mr. bush,  clinton, gop \\
   
    & \legendsquare{Legality} supreme court, justice, federal judge, court, board immigration appeal, high court, judge, justice department\\
   
    & \legendsquare{Public} organizer, activist, protester, demonstrator, marcher, immigrant advocate, mayor, poll, group, coalition\\
    
    & \legendsquare{Economic} grower, farmer, taxpayer, company, employer, native, foreign, business, high-tech company, federal government\\
    
    & \legendsquare{Morality} church, god, bishop, bible, course action, christ, parishioner, archbishop, organization, local jewish leader \\
    
    & \legendsquare{Fairness} Mr.Crocker's letter, deputization, immigrants' rights group, American, critic, student, bigotry, commission\\
    
    & \legendsquare{Policy} executive order, Trump administration, legislation, legislator, measure, Obama administration, bill, provision\\
    
    & \legendsquare{Crime} federal agent, prosecutor, investigator, federal authority, federal immigration agent, Tyson, federal prosecutor \\
    
    & \legendsquare{Security} FBI, ashcroft, coast guard, border patrol, Mr. Ashcroft, terrorist, homeland security official, border patrol agent\\
    
    & \legendsquare{Health} health official, federal health official, doctor, migrant, hospital, disease, smuggler, patient, virus, mother, novice\\
    
    & \legendsquare{Life} new office, teacher, perez, parent, mother, student, danilda, honduran immigrant christino castro, child, family\\
    
    & \legendsquare{Cultural} Ziegler, census bureau, foreign, child immigrant, people want citizen country, new immigrant, museum\\
    
    & \legendsquare{External} Israel, Cuba, Mexican government, bahamian government, Castro, Clinton administration, Mexico, Cuban\\
    
    \midrule
    

   \multirow{14}{*}{predicate} &
    \legendsquare{Capacity} process, handle, swamp, accommodate, wait, exceed, fill, reduce, flood, crowd, clear, jam, overwhelm, rush\\
   
    & \legendsquare{Political} veto, defeat, vote, win, introduce, endorse, elect, oppose, overhaul, stall, criticize, derail, unveil, push, split, do\\
   
    & \legendsquare{Legality} sue, uphold, entitle, appeal, shall, violate, file, rule, challenge, qualify, dismiss, prove, expire, pending, block\\
   
    & \legendsquare{Public} chant, march, protest, rally, wave, gather, organize, stag, shout, cheer, oppose, demand, denounce, favor, draw\\
    
    & \legendsquare{Economic} cost, import, invest, afford, contribute, save, employ, earn, cut, fill, fund, compete, depend, attract, educate, buy\\
    
    & \legendsquare{Morality} pray, worship, forgive, love, welcome, thank, bless, stand, honor, urge, hope, speak, join, recognize, offer\\
    
    & \legendsquare{Fairness} discriminate, treat, single, persecute, complain, deserve, harass, punish, offend, tolerate, target, ignore\\
    
    & \legendsquare{Policy} verify, prohibit, streamline, crack, aim, implement, propose, tighten, fix, require, overhaul, bar, introduce, ban\\
    
    & \legendsquare{Crime} indict, sentence, plead, convict, fine, conspire, smuggle, harbor, sell, acquit, raid, shoot, commit, nab, arrest\\
    
    & \legendsquare{Security} patrol, beef, track, apprehend, secure, tighten, overstay, intercept, investigate, link, cross, pose, deploy, fly\\
    
    & \legendsquare{Health} infect, injure, hospitalize, die, drown, suffer, rescue, kill, cross, fell, crash, treat, flip, scorch, test, cause, hit\\
    
    & \legendsquare{Life} cry, graduate, felt, imagine, feel, sleep, reunite, enrol, worry, enroll, sit, remember, escape, miss, love, learn\\
    
    & \legendsquare{Cultural} assimilate, settle, celebrate, immigrate, account, teach, welcome, learn, found, preserve, spangle, publish, melt\\
    
    & \legendsquare{External} discuss, resume, press, ease, accept, visit, defect, meet, express, elect, legalize, urge, agree, refuse, talk, promote\\
    
    \midrule
    

   \multirow{14}{*}{\argone} &
    \legendsquare{Capacity} application, foreign worker, applicant, cap, number, application process, time, appointment, green card, staff\\
    
    & \legendsquare{Political} amendment, reform, legislation, voter, senate bill, immigration bill, Latino voter, election, immigration law\\
   
    & \legendsquare{Legality} asylum, lawsuit, suit, status, case, Elian, legal status, permanent residency, license, decision, ruling, hearing\\
   
    & \legendsquare{Public} rally, marcher, march, protest, movement, crowd, event, attention, message, poll, protester, immigration reform\\
    
    & \legendsquare{Economic} wage, economy, tax, state tuition, money, job, cost, income, worker, service, business, fee, foreign, budget\\
    
    & \legendsquare{Morality} church, sanctuary, god, refuge, yoga, faith, campaign, politics, home, better life, violence, pray, change heart\\
    
    & \legendsquare{Fairness} racial profiling, discrimination, due process, right, Latino, every legal immigrant, Hispanic, advantage, woman\\
    
    & \legendsquare{Policy} path, system, legal immigration, legal status, immigration status, program, policy, require legislative approval\\
    
    & \legendsquare{Crime} guilty, crime, investigation, deportation proceeding, criminal, illegal alien, bribe, drug, trinket, death penalty\\
    
    & \legendsquare{Security} border security, security, terrorist, national security, terrorism, fence, wall, border, information, troop\\
    
    & \legendsquare{Health} medical care, disease, health insurance, health care, medical treatment, body, treatment, coverage, prenatal care\\
    
    & \legendsquare{Life} food stamp, English, high school, goodbye, better life, everything, life, school, family, poverty, kid, father\\
    
    & \legendsquare{Cultural} population, immigrant population, black, America, Asian, resident, Spanish, home, book, American dream\\
    
    & \legendsquare{External} Cuba, agreement, meeting, Mexico, Cuban, negotiation, office, Haiti, Mexican government, island, migrant\\

   \bottomrule
 \end{tabular}}}
 \caption{\label{tab:descriptors_supplementary} Spans inferred as most highly associated with: Capacity \& Resources (\legendsquare{Capacity}), Political (\legendsquare{Political}), Legality (\legendsquare{Legality}), Public Sentiment (\legendsquare{Public}), Economic (\legendsquare{Economic}), Morality (\legendsquare{Morality}), Fairness \& Equality (\legendsquare{Fairness}), Policy Prescription \& Evaluation (\legendsquare{Policy}), Crime \& Punishment (\legendsquare{Crime}),   Security \& Defense (\legendsquare{Security}), Health \& Safety (\legendsquare{Health}), Quality of Life (\legendsquare{Life}), Cultural Identity(\legendsquare{Cultural}), External Regulation 
 \& Reputation (\legendsquare{External}) frames.}
\end{table*}

\end{document}